\documentclass[runningheads]{llncs}
\usepackage[T1]{fontenc}
\usepackage{cite}
\usepackage{graphicx}
\usepackage{booktabs,makecell}
\usepackage{caption}
\usepackage{subcaption}

\usepackage{hyperref}
\usepackage{color}

\begin{document}
\title{Improving Knot Prediction in Wood Logs with Longitudinal Feature Propagation}
%
%
\newif\ifanonymous
\newif\ifonlyabstract

\anonymoustrue
\onlyabstractfalse

\author{
\ifanonymous 
Salim Khazem\inst{1,2}\orcidID{0000-0001-5958-6120}, Jeremy Fix\inst{2,4}\orcidID{0000-0003-1889-8886} \and C\'edric Pradalier\inst{1,3}\orcidID{0000-0002-1746-2733}
\fi
}

\authorrunning{S. Khazem et al.}

%
\institute{
\ifanonymous
GeorgiaTech-CNRS IRL 2958, Metz, France\and Centralesupelec, Metz, France \and GeorgiaTech Europe, Metz, France \and LORIA, CNRS UMR 7503, Metz, France\\ \email{salim.khazem@centralesupelec.fr}
\fi
}
\maketitle 
\begin{abstract}

The quality of a wood log in the wood industry depends heavily on the presence of both outer and inner defects, including inner knots that are a result of the growth of tree branches. Today, locating the inner knots require the use of expensive equipment such as X-ray scanners. In this paper, we address the task of predicting the location of inner defects from the outer shape of the logs. The dataset is built by extracting both the contours and the knots with X-ray measurements. We propose to solve this binary segmentation task by leveraging convolutional recurrent neural networks. Once the neural network is trained, inference can be performed from the outer shape measured with cheap devices such as laser profilers. We demonstrate the effectiveness of our approach on fir and spruce tree species and perform ablation on the recurrence to demonstrate its importance.

\end{abstract}

\ifonlyabstract
\else 

\section{Introduction}
Distribution of knots within logs is one of the most important factor in wood processing chain since it determines how the log will be sliced and used. A knot is defined as a piece of a branch that is lodged in a stem and often starts at the stem pith. Knots come in various dimensions, shapes and trajectories inside the trunk, these characteristics often depend on tree specie and environmental factors \cite{article1}. In wood processing, the knots are considered as defects that affect the quality of logs; hence, detecting their features such as position, size and angle of inclination are relevant and crucial for foresters and sawyers. Knowing these characteristics before the tree processing could generate a relative gain of 15-18\% in value of products \cite{bhandarkar_catalog_1999}. Nowadays, internal prediction of tree trunk density from bark observation is a complex and tedious task that requires a lot of human expertise or cannot be performed without expensive X-rays machines. In recent years, with the advent and success of deep learning, convolutional neural networks have achieved great performances on a variety of tasks such as object detection and image classification due to their strong features extraction capabilities\cite{Lecun2015,goodfellow2016deep}. Compared to traditional methods, the data driven deep learning based approaches learn discriminative characteristics from annotated data automatically instead of human engineering. While the era of deep learning led to significant improvements in several areas of computer vision and natural language processing, there are still only a few paper which study the interest of these approaches for forestry and wood processing industry. This is due to the lack of open data, but also due to the lack of transfer of architectures that have demonstrated their efficiency in computer vision to the specific tasks of the forestry and wood processing industry. In this paper, we propose to explore an original task that does not seem to bear resemblance with a task in another domain: predicting the inner structure from the outer appearance of a wood log. The internal knots of a wood log are a consequence of the growth of branches of the tree and there is therefore, at least for some species, a causality between the presence of an inner knot and the growth or scar of an external branch. As we will demonstrate in the paper, the deformation of the outer surface of the tree, which is the consequence of the presence of branches, allows inferring the location and shape of inner knots. Our experiments are carried on conifers for which there is a clear relationship between the growth of branch and the knots. However, for other species such as deciduous trees, this relationship is unclear, and the task remains challenging.

\begin{figure*}[htbp]
\vskip -0.2in
\begin{center}
\centerline{\includegraphics[width=0.85\columnwidth]{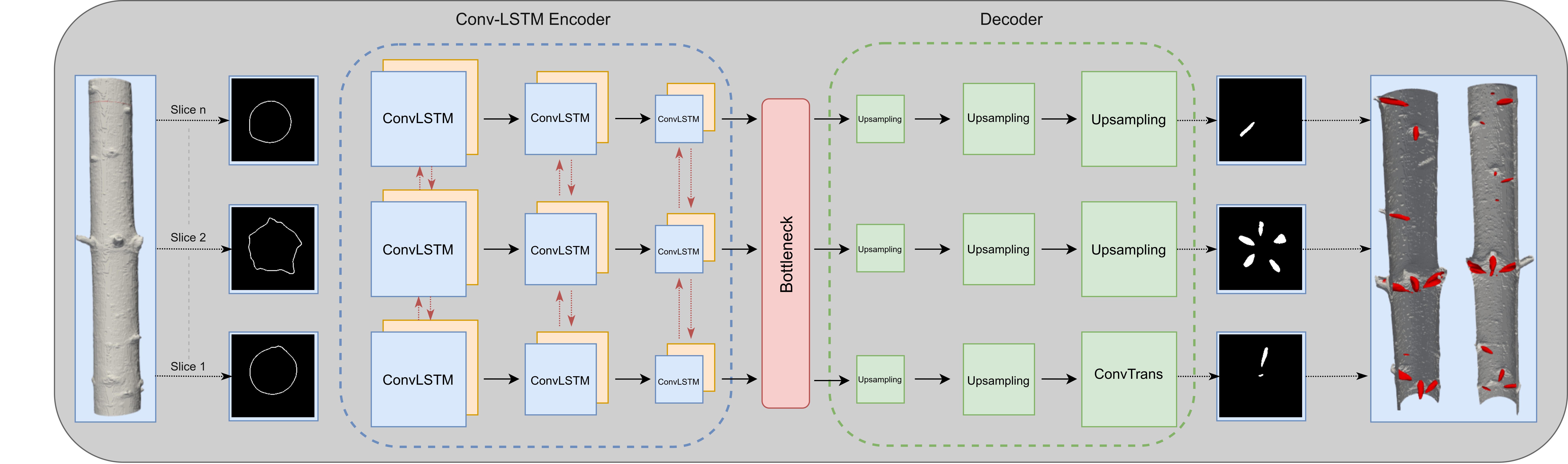}}
\caption{The recurrent neural network involves a recurrent encoder and feedforward decoder. The context along the slice dimension is propagated with convolutional LSTMs.}
\label{convlstm-network}
\end{center}
\vskip -0.5in
\end{figure*}

To solve the task of predicting the inner knots from the outer contour, we consider convolutional neural networks of the encoder-decoder family, where the encoder extracts features for the contour which are then used to decode the presence of a knot as a binary mask. Regularly spaced contour slices of the tree are provided as input to the network. As the presence of a knot is causally linked with a deformation of the contour due to a branch, inferring a knot needs to integrate features from the contour slices further away up or down the tree. To propagate these features between different slices, we consider convolutional LSTM, which are convolutional bidirectional recurrent neural networks\cite{shi2015convolutional}. A convolutional recurrent network keeps the spatial structure of the representation and extracts features along the recurrent paths by applying convolutions rather than dense matrix products. This has the benefit of reducing the cost of the network. In our task, this makes sense because a knot progressively diffuses within the wood as one moves along the longitudinal axis of the tree. That progressive diffusion induces that relevant features can be extracted locally, without having to resort to longer range interactions. Finally, given knots have various shapes and diffuses along a varying number of slices, using LSTMs lets the neural network learn how many slices need to be integrated to properly recover the shape of the knot. In summary, the main contribution of our work lies in two parts: 
\begin{itemize}
    \item we propose to address an original machine learning task that is also valuable for the forestry industry, namely, the prediction of inner defects given observations of the outer deformation of a wood log,
    \item we demonstrate the efficiency of integrating recurrent connections in the segmentation network to solve this task.
\end{itemize}

The code used for running all the experiments of this paper are available on the following github repository: \url{https://github.com/jeremyfix/icvs2023}.

\section{Related Work}

\textbf{Semantic segmentation} is a fundamental task in computer vision where the goal is to predict the label of each pixel in an image. Deep learning architectures for this task are typically based on the auto-encoder architecture. An autoencoder consists of an encoder and a decoder. The encoder maps the input data to a lower-dimensional latent space representation, while the decoder maps the latent space back to the original input data dimension  \cite{autoencoder}. In semantic segmentation, the decoder decodes the target labels instead of reconstructing the input. Fully Convolutional Networks (FCN)\cite{FCN_paper} is an important approach in semantic segmentation and has influenced the design of modern segmentation network. Other refinements of the encoder-decoder structure, such as U-Net and SegNet, have also been proposed in the literature \cite{Unet, SegNet}.

\textbf{Recurrent Neural Networks} have been introduced to deal with sequence data. They can learn the required size of the temporal window to gather the context required for taking a decision at any given time. The difficulty to integrate and propagate information through time, which is the foundation of the fundamental deep learning problem \cite{Hochreiter1991} of the vanishing/exploding gradient, has led authors to design dedicated memory units. Representatives of this family are the Long Short Term Memory networks (LSTMs)\cite{Hochreiter1997,Gers2000} and Gated Recurrent Units networks (GRUs)\cite{Cho2014}.

\textbf{Convolutional LSTM} preserves the convolutional nature of the data\cite{shi2015convolutional}. Indeed, the recurrent weights in the LSTMs involve dense connections and do not exploit the spatial structure of the data they process. Convolution LSTM, by considering convolutional recurrent ways, do preserve the spatial nature of data and reduces the number of parameters required in the recurrent connections. In the original paper, the convolutional LSTMs have been successfully applied to spatio-temporal sequences for weather forecasting.

In our work, we use an encoder-decoder architecture to predict the knot distribution (binary mask) from the slices of contours of the tree. To propagate encoder features through the slices, the encoder involves recurrent connections. In order to keep the convolutional nature of the representations, the encoder involves convolutional LSTM networks. Alternatively, we could have considered a 3D convolutional encoder, but this would have fixed the size of the slice context necessary to form a prediction. Using LSTMs let the network learn which contour features influence which other contour features.

\section{Methodology}

\subsection{Data Preprocessing}

In order to learn to transform the knot distribution from the contour of trees, we need aligned pairs of contours and knot masks. To build the input and target, we considered the pipelines of~\cite{khazem2023} for segmenting knots and identifying the contours by using X-rays data. Note that even though the pipelines of~\cite{khazem2023} are used to acquire data from X-rays images. The main objective of our approach is to avoid X-ray scanners and recover the external geometry from other modalities such as vision camera or laser profilers. The dataset is built from $27$ fir trees and $15$ spruce trees, with slices every $1.25$ mm for tree sections of 1 meter long in average, which makes a total of $30100$ slices. Each image is an $512\times 512$ that is downscaled to $256 \times 256$ for the extraction of the contour and knot segmentation, and further downscaled to $192 \times 192$ for the sequence models presented in this paper. Every tree is sliced in blocks of $40$ consecutive slices. In the following of the paper, the axis along which the slices are stacked will be referred as either the longitudinal axis or the z-axis for short. In the experiments, we used 18 fir tree and 8 spruce tree for the training set, we used 4 fir tree and 
2 spruce tree for the validation and 5 tree of each specie for the test set. Note that, each tree is represented with by 800 slices.

\subsection{Neural network architectures without recurrent connections\label{subsec:ffnseg}}

We trained two feedforward neural networks based on U-Net\cite{Unet} and SegNet\cite{SegNet} in order to obtain a baseline to compare with the architecture involving recurrent connections along the z-axis. Although the U-Net and SegNet do not involve recurrent connections, these have been trained on the same data as the recurrent networks, e.g., stacks of slices. This allows to guarantee that training has been performed on the same data and the metrics are computed the same way. The U-Net encoder involves fewer channels than the original network to fit with the input data. The upsampling along the decoder path is performed using a nearest-pixel policy. Along the decoding path, the encoder features are concatenated with the decoder features. The SegNet encoder involves less channels and convolutional layers than the original network. The number of blocks and channels is reduced with respect to the original SegNet because our inputs are smaller.

\subsection{Neural network architectures with recurrent connections}

In order to propagate the contextual features of the contours in the encoder, we also consider neural network architectures with recurrent connections along the slice dimension (longitudinal axis of the tree). Recurrent connections are implemented with convolutional LSTMs which allow the network to learn which slice is impacting the features of another slice. We remind that the knots within a log can be causally linked to the presence of a branch. Instead of a fully connected LSTM, the convolutional LSTM involves fewer parameters by exploiting the spatial structure of the input data. In this paper, we consider recurrent connections only in the encoder part and not in the decoder part. The rationale is that introducing recurrent connections in the encoder allows the network to propagate contour features through the slices, and our experiments show that this is already sufficient to get good performances. These recurrent connections are bidirectional to allow information to propagate in both directions along the longitudinal axis. For the decoder, we do not add recurrent connections. That could be helpful but at a higher computational cost, and our experiments already demonstrated good performances with recurrent connections only in the encoder. The neural network architecture is depicted on Figure~\ref{convlstm-network}.

The recurrent encoder is built from $3$ consecutive ConvLSTMs bidirectional blocks. Every block has the same number of memory cells than the size of the spatial dimensions times the channel dimension. The input, output, and forget gates compute their values from a convolution with kernel size $3$ from the “previous” sequence index (here, previous is to be considered along the longitudinal z-axis and be either following upward or downward directions given we consider bidirectional LSTMs). We use the same representation depth than for the SegNet with $32$, $48$ and $64$ channels and a maxpooling layer is placed after every ConvLSTM layer to downscale spatially the representation by a factor of $2$. The decoder is not recurrent and is the same as for our SegNet, namely $3$ consecutive blocks with an upsampling (nearest) followed by a $2\times[Conv2D(3\times 3)-BatchNorm-ReLU]$ block. The final layer is a $Conv(1\times 1)$ to output the unnormalized scores for the classification of every pixel.

\subsection{Evaluation metrics}
Our experiments use different quantitative metrics to evaluate the quality and the performance of our method. For the segmentation task, the ground truth output is usually very sparse and there are much more negatives than positives. Hence, we need to use evaluation metrics that are not biased due to this class imbalance. We used the Dice similarity coefficient (Dice) \cite{dice_score}, which is also known as F1-score as overlap metric, the Hausdorff Distance (HD) \cite{hausdorff_distance} as distance-based metric, and the Cohen’s Kappa $\kappa$ \cite{cohen1960coefficient, muller2022towards} as counting-based metric to evaluate the segmentation results. 

The Hausdorff distance complements the Dice similarity because it indicates if false positives are close to a patch of positives or further away, while the Cohen's Kappa indicates the agreement between ground truth and the prediction. For each pixel, Cohen's Kappa compares the labels assigned by the model with the ground truth and measures the degree of agreement between them. The Cohen's Kappa ranges from -1 to 1 where a value of $1$ indicates perfect agreement between the prediction and ground truth, whereas $0$ indicates a prediction which is not better than random guessing and a negative value indicates less agreement than expected by chance. The advantage of using Cohen's Kappa is that it takes into account the possibility of chance agreement and provides a more accurate measure of agreement between prediction and ground truth, this is important in cases where the number of pixels assigned to each class is imbalanced. 

For the different equations, we denote FN, FP, TP, TN respectively the number of false negatives, false positives, true positives and true negatives, where  $\hat{y}$ is defined as final prediction computed by thresholding the output probability computed by the network (the threshold is set to $0.5$ for all the experiments), and $y$ the true value to be predicted (a mask, made of either $1$ for a pixel belonging to a knot, or $0$ otherwise). The metrics are always evaluated on the whole volume of $40$ slices. As mentioned in section~\ref{subsec:ffnseg}, even the feedforward neural networks (SegNet and UNet) are trained on the volumes. Although these networks do not propagate informations throught the longitudinal axis, training and evaluating these networks on the volume allow to have comparable measures (averaged on the same data). The value of the Hausdorff Distance is reported in millimeters. The metrics reported in the result section are averaged over the total number of volumes in the considered fold.

\newcommand{\myimgsize}{0.13}
\newcommand{\myraiseheight}{15pt}

\begin{table}
\begin{minipage}{0.4\linewidth}
\begin{tabular}{l||ccr}
Method  & Dice/F1 $\uparrow$ &  HD $\downarrow$ \\
\hline\hline
SegNet  & 0.68 & 26.18\\
U-Net  & 0.72 & 47.80\\\hline
ConvLSTM & \textbf{0.84} & \textbf{17.34}\\
\end{tabular}
\end{minipage}\hspace{0.05\columnwidth}
\begin{minipage}{0.6\linewidth}
    \setlength\tabcolsep{1pt}
    \renewcommand{\arraystretch}{0.3}
    \begin{tabular}{cccccc}
    \raisebox{\myraiseheight}{\parbox[b]{.03\linewidth}{\rotatebox[origin=c]{90}{Input}}}
    &\includegraphics[width=\myimgsize\columnwidth]{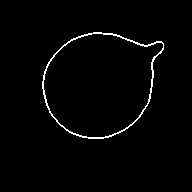} & \includegraphics[width=\myimgsize\columnwidth]{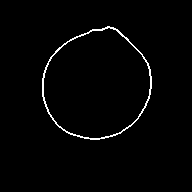} &
    \includegraphics[width=\myimgsize\columnwidth]{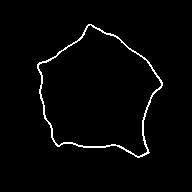} &
    \includegraphics[width=\myimgsize\columnwidth]{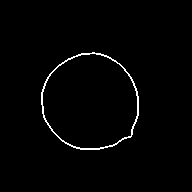} &
    \includegraphics[width=\myimgsize\columnwidth]{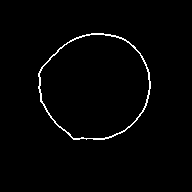} \\
    \raisebox{\myraiseheight}{\parbox[b]{.03\linewidth}{\rotatebox[origin=c]{90}{GT}}}
    &\includegraphics[width=\myimgsize\columnwidth]{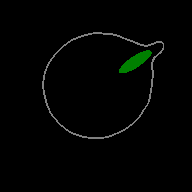} & \includegraphics[width=\myimgsize\columnwidth]{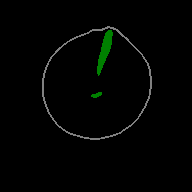} &
    \includegraphics[width=\myimgsize\columnwidth]{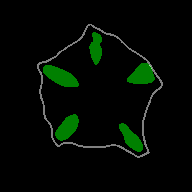} &
    \includegraphics[width=\myimgsize\columnwidth]{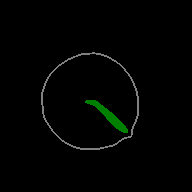} &
    \includegraphics[width=\myimgsize\columnwidth]{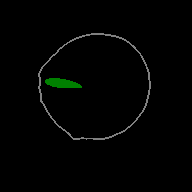} \\
    \raisebox{\myraiseheight}{\parbox[b]{.03\linewidth}{\rotatebox[origin=c]{90}{SegNet}}}
    &\includegraphics[width=\myimgsize\columnwidth]{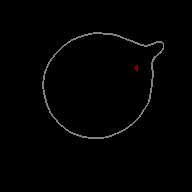} & \includegraphics[width=\myimgsize\columnwidth]{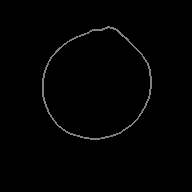} &
    \includegraphics[width=\myimgsize\columnwidth]{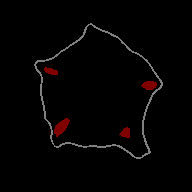} &
    \includegraphics[width=\myimgsize\columnwidth]{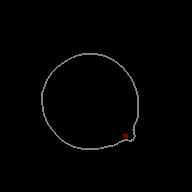} &
    \includegraphics[width=\myimgsize\columnwidth]{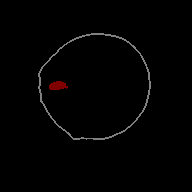} \\
    \raisebox{\myraiseheight}{\parbox[b]{.03\linewidth}{\rotatebox[origin=c]{90}{CLSTM}}}
    &\includegraphics[width=\myimgsize\columnwidth]{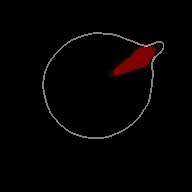} & \includegraphics[width=\myimgsize\columnwidth]{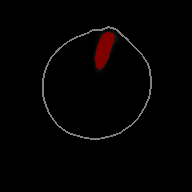} &
    \includegraphics[width=\myimgsize\columnwidth]{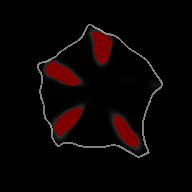} &
    \includegraphics[width=\myimgsize\columnwidth]{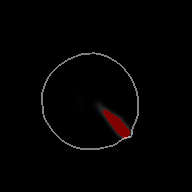} &
    \includegraphics[width=\myimgsize\columnwidth]{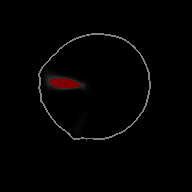} 
    \end{tabular}
\end{minipage}
\caption{Left) Comparison of the segmentation methods on Dice score and HD using the validation fold. Right) Results of the SegNet and ConvLSTM models for a Fir tree specie. The first row corresponds to the input images, the second row is the associated ground truth and the bottom ones are the predictions. These samples all belong to the validation fold. Every column corresponds to one of $5$ slices from different volumes.\label{tab:validation_metrics_and_samples}}
\end{table}

\subsection{Other experimental hyperparameters}
For all the experiments presented in the paper, the optimization followed the same schedule. The networks have been trained for $150$ epochs with a batch size of either $10$ for U-Nets and ConvLSTMs, reduced to $4$ for SegNet. The parameters have been optimized with Adam \cite{adam}, a base learning rate of $0.0001$. The loss is the binary cross entropy. ConvLSTMs trained for one week, the U-Net and SegNet trained for almost $10$ days, using two RTX 3090. The experiments were coded either with Tensorflow 2.4 \footnote{https://www.tensorflow.org} or Pytorch 1.9. We used Tensorboard\footnote{https://www.tensorflow.org/tensorboard} to track the experiments and log the curves (loss and the different metrics). For regularizing the ConvLSTM encoder-decoder, a dropout layer is inserted between the encoder and decoder parts with a probability of $10\%$ to mask a neuron. Following the original papers of U-Net and SegNet, we did not insert dropout layers in these networks. In all the trainings, data augmentation is applied to the input data with a random rotation out of $8$ possible angles, and horizontal flip with a probability of $0.5$.

\section{Results}

In this section, we present both quantitatively and qualitatively the performances of the various models on the prediction of knots.  The results on the validation fold and test folds are provided respectively in table~\ref{tab:validation_metrics_and_samples} and table~\ref{tab:test_metrics}. 

For all the metrics, the ConvLSTM model performs better than the neural networks without recurrent connections. Looking only at the DICE and HD metrics, it seems that even without the recurrent connections, both the SegNet and U-Net perform reasonably well on the task. However, we observed qualitatively that this is not really the case as several knots are not predicted by these models. In that respect, the kappa metric seems to reflect more the difference in performance between the feedforward and recurrent networks. 

Including the context with the recurrent connections in the encoder provides a boost in performance. The quality of the segmentation of the recurrent network is better if we look at the Hausdorff distance, which means that the predicted masks with the ConvLSTM are closer in distance to the ground truth than with the non-recurrent segmentation networks. The Hausdorff distance is given in millimeters, and we remind that the slices are $192\times 192$ pixels which correspond to $192\mbox{mm} \times 192\mbox{mm}$. Additionally, we computed on the test set the Cohen's Kappa to evaluate the agreement between the predicted masks and the ground truth. The results show that the ConvLSTM achieves a score of $0.41$ for fir trees and $0.21$ for spruce
indicating respectively moderate agreement and fair agreement, while the non-recurrent networks score lower with Kappa values between 0.05 and 0.12 for both species indicating very weak agreement. These findings demonstrate the boost provided by the recurrent networks.

\begin{table}
\begin{minipage}{0.45\linewidth}
\begin{tabular}{lc||cccr}
Method & Specie & Dice $\uparrow$ & HD $\downarrow$ & Kappa $\uparrow$ \\
\hline\hline
SegNet & Fir    & 0.68 & 17.04& 0.12\\
SegNet & Spruce & 0.67 & 33.14& 0.06\\
U-Net & Fir    & 0.69 & 37.95& 0.10\\
U-Net & Spruce & 0.68 & 56.62& 0.05\\\hline
ConvLSTM & Fir    & \textbf{0.74} & \textbf{12.68}& \textbf{0.41}\\
ConvLSTM & Spruce & \textbf{0.70} & \textbf{22.11}& \textbf{0.21} \\
\end{tabular}
\end{minipage}\hspace{0.1\columnwidth}
\begin{minipage}{0.45\linewidth}
\begin{tabular}{c | c || c c c}\hline
     Specie& Tree & \multicolumn{3}{c}{Metrics}\\
    &ID& Dice $\uparrow$& HD $\downarrow$ & Kappa $\uparrow$\\\hline    
     &4392&0.72&14.6 & 0.28 \\
     &4394&0.75&16.3 & 0.29\\
    \textbf{Fir} &4396&0.78&8.0& 0.52\\
     &5027&0.84&6.5 & 0.50\\   
     &5028&0.78&8.4 & 0.53\\
     \hline 
     &4327&0.70&29.0& 0.12\\
     &4328&0.72&19.2&0.12\\
    \textbf{Spruce} &4329&0.73&9.1&0.25\\
     &4948&0.70&31.0& 0.11\\
     &4990&0.73&13.6&0.26\\
     \hline  
\end{tabular}
\end{minipage}
\caption{Left) Comparison of the segmentation methods on Dice, HD and Kappa metrics on the test fold. Right) Quantitative results of the ConvLSTM model for the different trees of the test set. These are the same trees than the ones used for table on the left. The metrics are averaged over all the volumes of the same tree. All the trees had almost the same number of slices (from $800$ to $810$ slices).\label{tab:test_metrics}}
\end{table}

In table~\ref{tab:test_metrics}, right, we provide the metrics of the ConvLSTM model on the different trees of the test fold, either fir or spruce. The metrics computed on individual trees are consistent with the averaged metrics computed over all the volumes and reported in table~\ref{tab:test_metrics}, left. However, some spruce trees are particularly challenging. That's the case for example for the trees $4327$ and $4948$ which have a really unconventional contours, strongly distorted for some reason unknown to the authors. This out-of-distribution contours probably explains why the model fails to correctly predict all the knots. In addition to these averaged metrics, we provide in Figure~\ref{fig:violinplot} the distribution of Cohen's Kappa metric computed on the test fold for both fir and spruce trees, for both the ConvLSTM and SegNet networks. We observe that the ConvLSTM model outperforms the SegNet for all trees for both species, with a clear separation between the distributions. Specifically, the ConvLSTM model achieves nearly a twofold improvement over the SegNet for almost all trees.


As the SegNet performs better on the test set than the U-Net, further comparison will only be made between SegNet and the ConvLSTM network. To better appreciate the difference in segmentation quality between the SegNet and ConvLSTM networks, the prediction masks of both networks on individual slices from different volumes are given in Table~\ref{tab:validation_metrics_and_samples}, right. On this figure, every column is a slice from a different volume of a fir tree and consecutive rows represent the input contour, the ground truth, the prediction of SegNet and the prediction of the ConvLSTM. From these $5$ samples, it appears that SegNet usually underestimates knots and sometimes, knots may be even not predicted at all. For the ConvLSTM, most knots are predicted, although the knots might be overestimated in shape.

\renewcommand{\myimgsize}{0.085}
\renewcommand{\myraiseheight}{15pt}
\begin{figure}[ht!]
    \setlength\tabcolsep{1pt}
    \renewcommand{\arraystretch}{0.3}
    \begin{tabular}{cccccc}
    \raisebox{\myraiseheight}{\parbox[b]{.03\linewidth}{\rotatebox[origin=c]{90}{Input}}}
    &\includegraphics[width=\myimgsize\columnwidth]{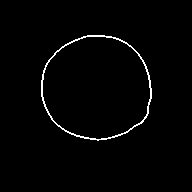} & \includegraphics[width=\myimgsize\columnwidth]{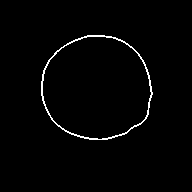} &
    \includegraphics[width=\myimgsize\columnwidth]{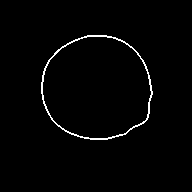} &
    \includegraphics[width=\myimgsize\columnwidth]{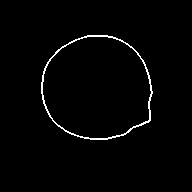} &
    \includegraphics[width=\myimgsize\columnwidth]{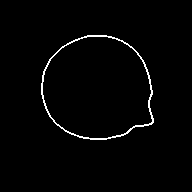} \\
    \raisebox{\myraiseheight}{\parbox[b]{.03\linewidth}{\rotatebox[origin=c]{90}{GT}}}
    &\includegraphics[width=\myimgsize\columnwidth]{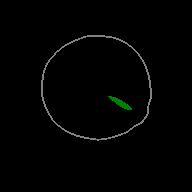} & \includegraphics[width=\myimgsize\columnwidth]{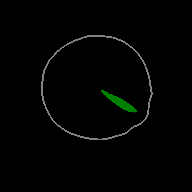} &
    \includegraphics[width=\myimgsize\columnwidth]{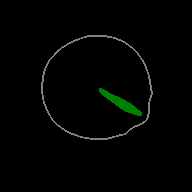} &
    \includegraphics[width=\myimgsize\columnwidth]{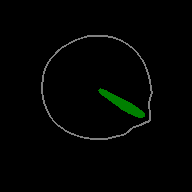} &
    \includegraphics[width=\myimgsize\columnwidth]{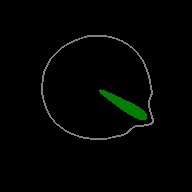} \\
    
    \raisebox{\myraiseheight}{\parbox[b]{.03\linewidth}{\rotatebox[origin=c]{90}{SegNet}}}
    &\includegraphics[width=\myimgsize\columnwidth]{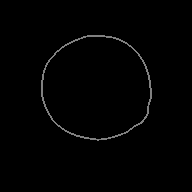} & \includegraphics[width=\myimgsize\columnwidth]{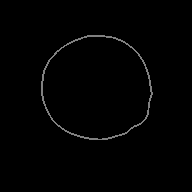} &
    \includegraphics[width=\myimgsize\columnwidth]{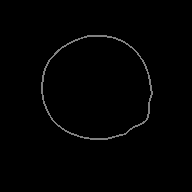} &
    \includegraphics[width=\myimgsize\columnwidth]{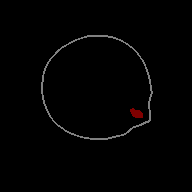} &
    \includegraphics[width=\myimgsize\columnwidth]{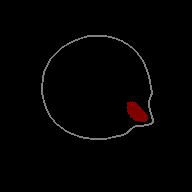} \\
    
    \raisebox{\myraiseheight}{\parbox[b]{.03\linewidth}{\rotatebox[origin=c]{90}{CLSTM}}}
    &\includegraphics[width=\myimgsize\columnwidth]{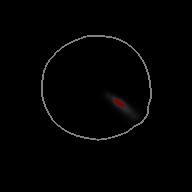} & \includegraphics[width=\myimgsize\columnwidth]{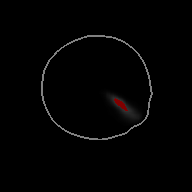} &
    \includegraphics[width=\myimgsize\columnwidth]{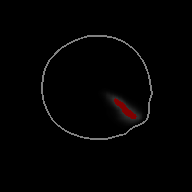} &
    \includegraphics[width=\myimgsize\columnwidth]{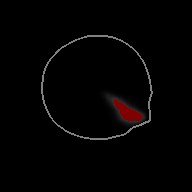} &
    \includegraphics[width=\myimgsize\columnwidth]{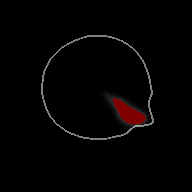} \\[0.1cm]
    \end{tabular}        
    \begin{tabular}{cccccc}
    \raisebox{\myraiseheight}{\parbox[b]{.03\linewidth}{\rotatebox[origin=c]{90}{Input}}}
    &\includegraphics[width=\myimgsize\columnwidth]{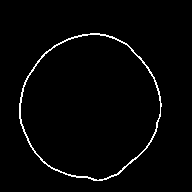} & \includegraphics[width=\myimgsize\columnwidth]{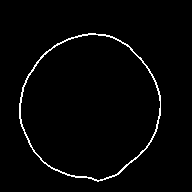} &
    \includegraphics[width=\myimgsize\columnwidth]{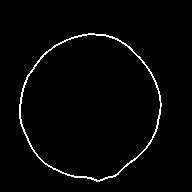} &
    \includegraphics[width=\myimgsize\columnwidth]{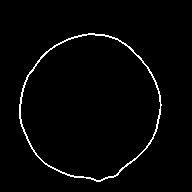} &
    \includegraphics[width=\myimgsize\columnwidth]{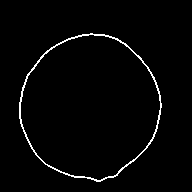} \\
    \raisebox{\myraiseheight}{\parbox[b]{.03\linewidth}{\rotatebox[origin=c]{90}{GT}}}
    &\includegraphics[width=\myimgsize\columnwidth]{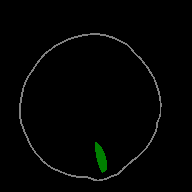} & \includegraphics[width=\myimgsize\columnwidth]{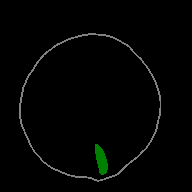} &
    \includegraphics[width=\myimgsize\columnwidth]{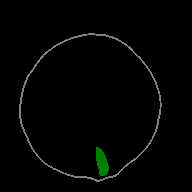} &
    \includegraphics[width=\myimgsize\columnwidth]{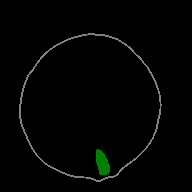} &
    \includegraphics[width=\myimgsize\columnwidth]{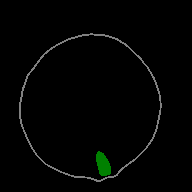} \\
    \raisebox{\myraiseheight}{\parbox[b]{.03\linewidth}{\rotatebox[origin=c]{90}{SegNet}}}
    &\includegraphics[width=\myimgsize\columnwidth]{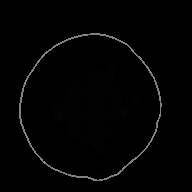} & \includegraphics[width=\myimgsize\columnwidth]{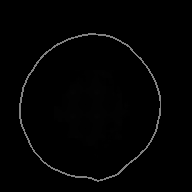} &
    \includegraphics[width=\myimgsize\columnwidth]{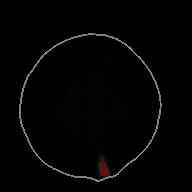} &
    \includegraphics[width=\myimgsize\columnwidth]{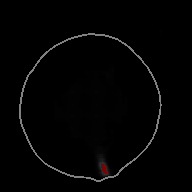} &
    \includegraphics[width=\myimgsize\columnwidth]{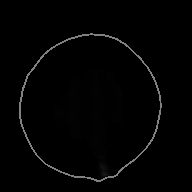} \\
    \raisebox{\myraiseheight}{\parbox[b]{.03\linewidth}{\rotatebox[origin=c]{90}{CLSTM}}}
    &\includegraphics[width=\myimgsize\columnwidth]{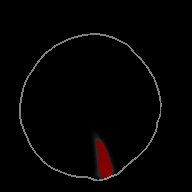} & \includegraphics[width=\myimgsize\columnwidth]{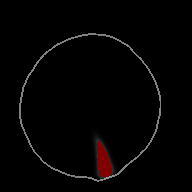} &
    \includegraphics[width=\myimgsize\columnwidth]{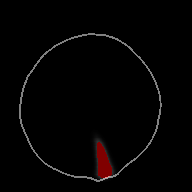} &
    \includegraphics[width=\myimgsize\columnwidth]{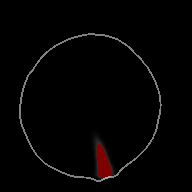} &
    \includegraphics[width=\myimgsize\columnwidth]{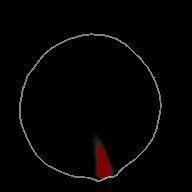} 
    \end{tabular}
    \caption{Results of the SegNet and ConvLSTM models for a fir tree (left) or spruce tree specie (right) on $5$ consecutive slices from the same volume. The first row corresponds to the input contours, the second row is the associated ground truth, and the two bottom rows are the predictions. These slices belong to a tree from the test set.}
    \label{fig:epiceq_Contours}
\end{figure}

The predictions on some consecutive slices of the same volume of a tree are shown on Figures~\ref{fig:epiceq_Contours} for respectively a fir tree and a spruce tree. On the fir tree (left), we see part of the branch getting out from the tree, which is the anchor feature from which a network could learn the presence of an inner knot. Indeed, the ConvLSTM seems to be able to propagate information through the slices with its recurrent connections, as it is able to predict the location of a knot on the first of the five slices. It seems unlikely a network could predict the presence of a knot solely based on the first slice, given the deformation of the latter is barely visible on the contour of this first slice.

\begin{figure}[ht!]
\includegraphics[width=0.50\textwidth]{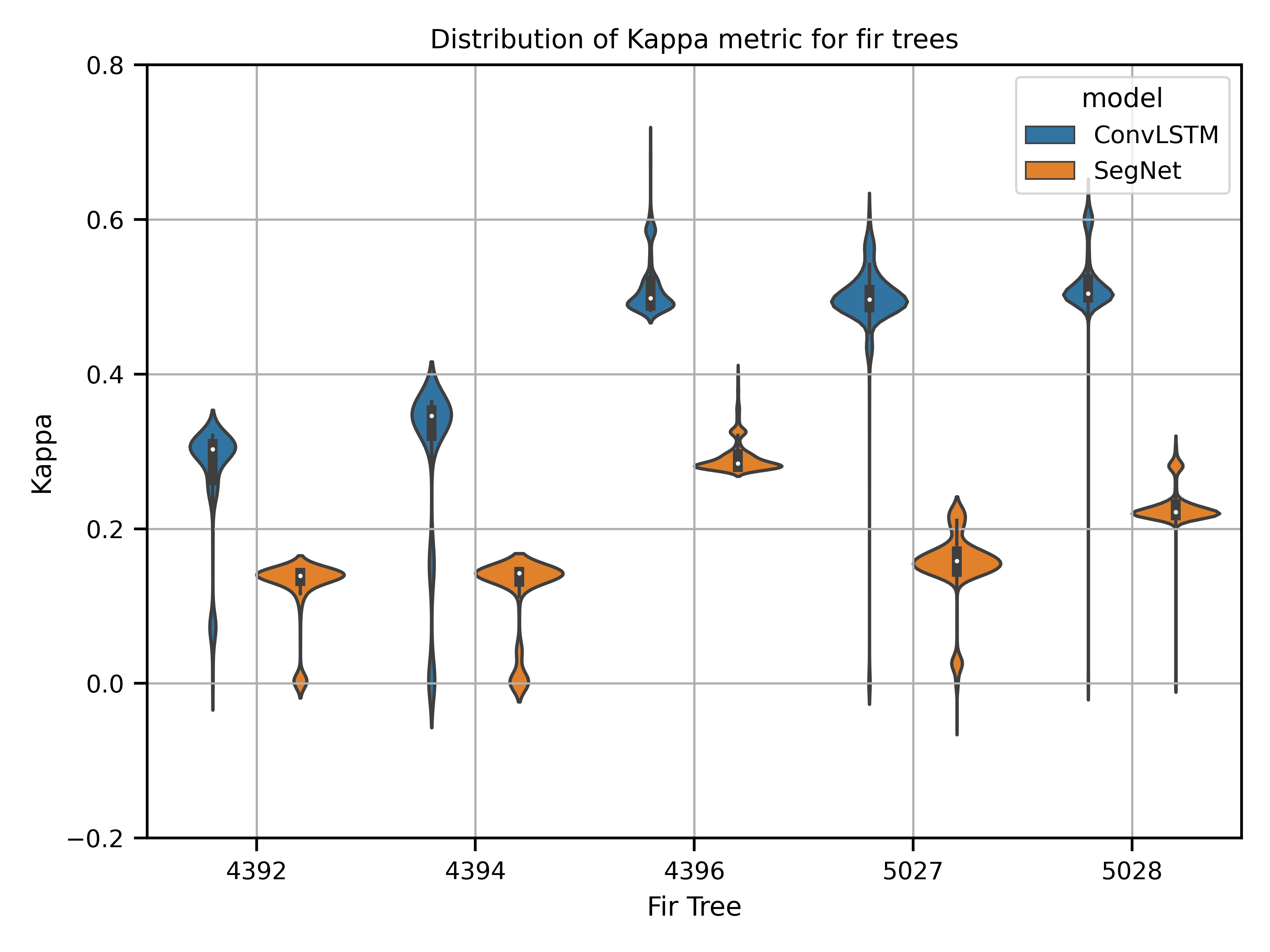} 
\includegraphics[width=0.50\textwidth]{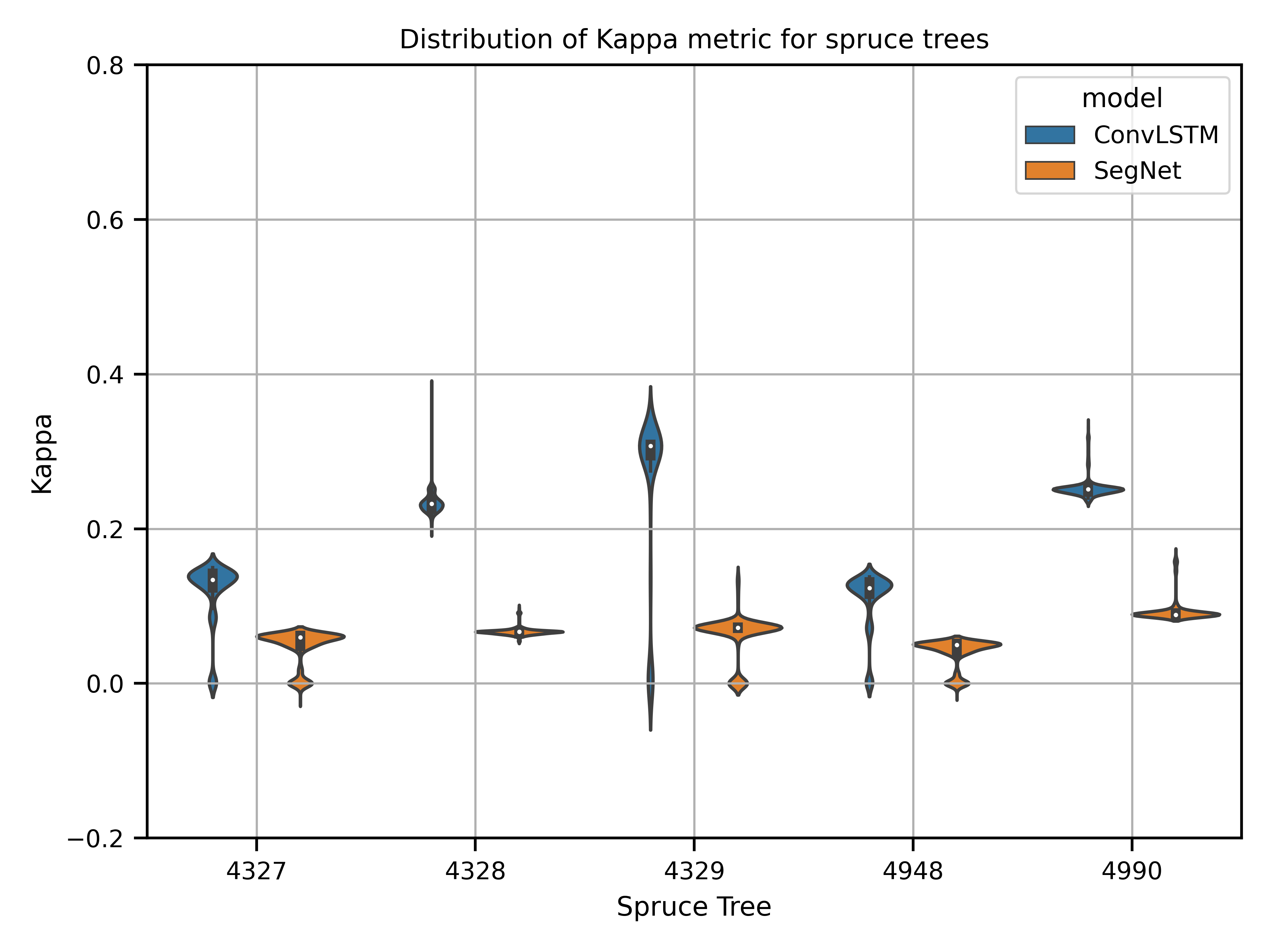} 
\caption{\label{fig:violinplot} Distribution of the Kappa metric on the fir and spruce trees of the test fold for both the SegNet and ConvLSTM neural networks. These are the same trees than used in table~\ref{tab:test_metrics}.}
\end{figure}

\begin{figure}[htbp]
\begin{center}
\begin{tabular}{cc}
\includegraphics[width=0.4\columnwidth]{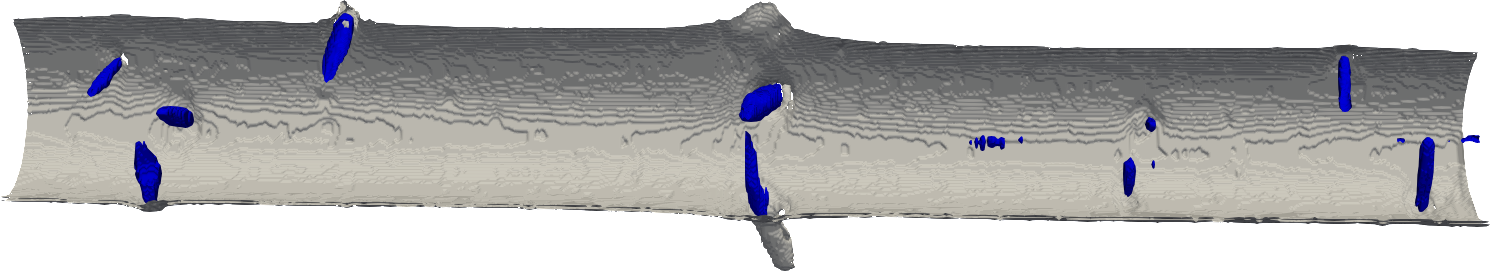} &
\includegraphics[width=0.4\columnwidth]{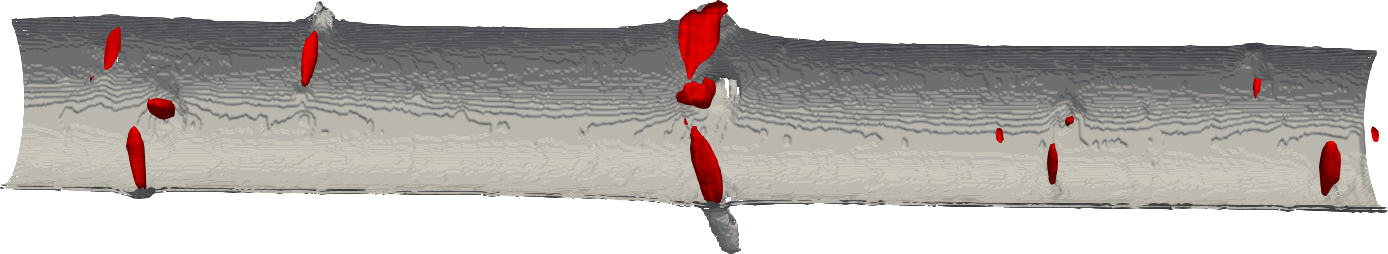}\\
\includegraphics[width=0.4\columnwidth]{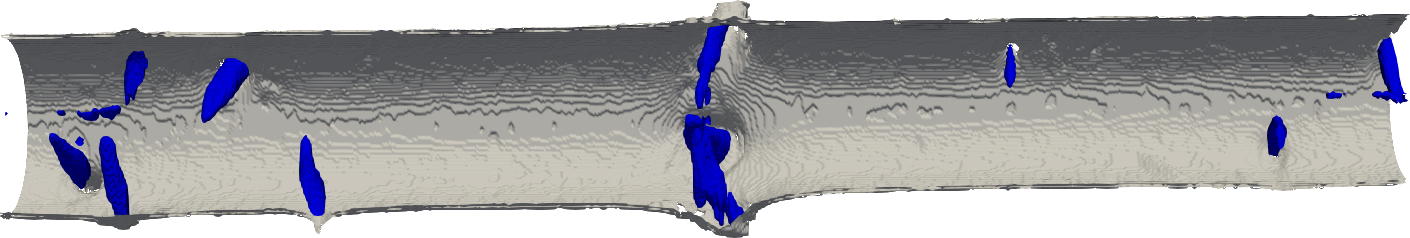} &
\includegraphics[width=0.4\columnwidth]{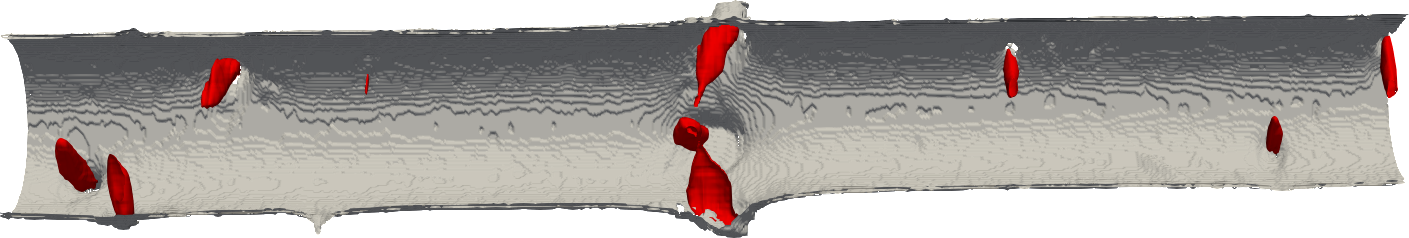}
\end{tabular}
\end{center}
\caption{\label{fig:volume3d} 3D representation of the ground truth (left) and prediction of the ConvLSTM (right) viewed from the side of the tree or the top on both sides. Generated with Paraview.}
\end{figure}

The figure~\ref{fig:volume3d} shows a 3D representation of the contour of a fir tree from the test set, as well as the ground truth and the prediction produced by the ConvLSTM. The full tree represents a set of $803$ slices, and all these slices are processed by sequences of $40$ slices, with a stride of $1$. From this full tree 3d representation, we observe that every knot present in the ground truth is also predicted by the ConvLSTM. It seems also that some knots may not have been correctly labeled as knots in the ground truth. This 3D representation also highlights the consistency of the knot predictions. From this representation, we also better see that there are various types of branch scars, some being clearly visible while others are more like little bumps on the surface of the tree. The smallest scars are certainly the ones for which it is the most challenging for the network to infer the location of knots, but even in some of these difficult cases, we see that the ConvLSTM model succeeds in predicting knots.

\section{Discussion}

In this paper, we investigated a machine learning task that is highly valuable for the forestry industry : predicting the location of inner defects, knots, from the outside appearance of the tree. From the machine learning perspective, this task is original. We addressed this problem by training various neural network architectures of the encoder/decoder family and the most promising tested architectures are the convolutional LSTMs which benefit from recurrent connections along the longitudinal axis of the tree to propagate contour features reflecting the scars of a branch to the slices where the knots must be predicted. Although from the averaged metrics (DICE and Hausdorff), the feedforward networks (SegNet, U-Net) seem to perform well, it turns out that their predictions are pretty bad when we observe them qualitatively. This is not the case for the convolutional LSTM model, which have better metrics and clearly better segmentation of the knots when we check them visually. This discrepancy needs further investigation, and it is unclear why good classification metrics would lead to bad segmentation. The performances of the networks appear to be more contrasted by the Cohen's Kappa. 

The data used by the proposed machine learning pipeline relies on the work of \cite{khazem2023} that extract contour and inner knots of tree logs from X-ray scans. X-ray scans are only essential to produce the targets but are not used by our proposed approach. The required contours for the model can be obtained using laser scanners. We have a work in progress to create a platform with calibrated lasers to extract the contour of a tree log. From a machine learning perspective, the input contours are sparse, but dense representations are used for encoding. There is room for improvement in encoding and decoding methods. Instead of using a binary mask, encoding the contour as a polygon and utilizing graph neural networks for differentiable feature learning could be more efficient. Furthermore, recent research on neural radiance fields \cite{Mildenhall2020} suggests the possibility of encoding a 3D volume as a parameterized function, eliminating the need to explicitly construct the 3D volume of knots. Although these ideas require experimentation, a lightweight recurrent encoding of contours that parameterizes a 3D knot density function holds promise.

\section*{Acknowledgment}
This research was made possible with the support from the French National Research Agency, in the framework of the project WoodSeer, ANR-19-CE10-011.

%
%
%


\bibliographystyle{splncs04}
\bibliography{references}

\begin{thebibliography}{10}
\providecommand{\url}[1]{\texttt{#1}}
\providecommand{\urlprefix}{URL }
\providecommand{\doi}[1]{https://doi.org/#1}

\bibitem{SegNet}
Badrinarayanan, V., Kendall, A., Cipolla, R.: Segnet: A deep convolutional
  encoder-decoder architecture for image segmentation. IEEE transactions on
  pattern analysis and machine intelligence  \textbf{39}(12),  2481--2495
  (2017)

\bibitem{bhandarkar_catalog_1999}
Bhandarkar, S.M., Faust, T.D., Tang, M.: {CATALOG}: a system for detection and
  rendering of internal log defects using computer tomography. Machine Vision
  and Applications  \textbf{11}(4),  171--190 (Dec 1999)

\bibitem{Cho2014}
Cho, K., van Merrienboer, B., Gulcehre, C., Bahdanau, D., Bougares, F.,
  Schwenk, H., Bengio, Y.: Learning {Phrase} {Representations} using {RNN}
  {Encoder}-{Decoder} for {Statistical} {Machine} {Translation}.
  arXiv:1406.1078 [cs, stat]  (Sep 2014)

\bibitem{cohen1960coefficient}
Cohen, J.: A coefficient of agreement for nominal scales. Educational and
  psychological measurement  \textbf{20}(1),  37--46 (1960)

\bibitem{dice_score}
Dice, L.R.: Measures of the amount of ecologic association between species.
  Ecology  \textbf{26}(3),  297--302 (1945)

\bibitem{Gers2000}
Gers, F.A., Schmidhuber, J.A., Cummins, F.A.: Learning to forget: Continual
  prediction with lstm. Neural Comput.  \textbf{12}(10),  2451--2471 (Oct 2000)

\bibitem{goodfellow2016deep}
Goodfellow, I., Bengio, Y., Courville, A.: Deep learning. MIT press (2016)

\bibitem{Hochreiter1997}
Hochreiter, S., Schmidhuber, J.: Long short-term memory. Neural Computation
  \textbf{9}(8),  1735--1780 (1997)

\bibitem{hausdorff_distance}
Huttenlocher, D.P., Klanderman, G.A., Rucklidge, W.J.: Comparing images using
  the hausdorff distance. IEEE Transactions on pattern analysis and machine
  intelligence  \textbf{15}(9),  850--863 (1993)

\bibitem{Hochreiter1991}
{Josef Hochreiter}: Untersuchungen zu dynamischen neuronalen {Netzen}. Ph.D.
  thesis, Technische Universität München (1991)

\bibitem{khazem2023}
Khazem, S., Richard, A., Fix, J., Pradalier, C.: Deep learning for the
  detection of semantic features in tree x-ray ct scans. Artificial
  Intelligence in Agriculture  (2023)

\bibitem{adam}
Kingma, D.P., Ba, J.: Adam: A method for stochastic optimization. arXiv
  preprint arXiv:1412.6980  (2014)

\bibitem{Lecun2015}
LeCun, Y., Bengio, Y., Hinton, G.: Deep learning. nature  \textbf{521}(7553),
  ~436 (2015)

\bibitem{FCN_paper}
Long, J., Shelhamer, E., Darrell, T.: Fully convolutional networks for semantic
  segmentation. In: Proceedings of the IEEE conference on computer vision and
  pattern recognition. pp. 3431--3440 (2015)

\bibitem{article1}
Longo, B., Br\"{u}chert, F., Becker, G., Sauter, U.: Validation of a {CT} knot
  detection algorithm on fresh {D}ouglas-fir ({P}seudotsuga menziesii ({M}irb.)
  {F}ranco) logs. Annals of Forest Science  \textbf{76} (06 2019)

\bibitem{Mildenhall2020}
Mildenhall, B., Srinivasan, P.P., Tancik, M., Barron, J.T., Ramamoorthi, R.,
  Ng, R.: Nerf: Representing scenes as neural radiance fields for view
  synthesis. In: ECCV (2020)

\bibitem{muller2022towards}
M{\"u}ller, D., Soto-Rey, I., Kramer, F.: Towards a guideline for evaluation
  metrics in medical image segmentation. BMC Research Notes  \textbf{15}(1),
  ~1--8 (2022)

\bibitem{Unet}
Ronneberger, O., Fischer, P., Brox, T.: U-net: Convolutional networks for
  biomedical image segmentation. In: International Conference on Medical image
  computing and computer-assisted intervention. pp. 234--241. Springer (2015)

\bibitem{shi2015convolutional}
Shi, X., Chen, Z., Wang, H., Yeung, D.Y., Wong, W.K., Woo, W.c.: Convolutional
  lstm network: A machine learning approach for precipitation nowcasting.
  Advances in neural information processing systems  \textbf{28} (2015)

\bibitem{autoencoder}
Yu, S., Principe, J.C.: Understanding autoencoders with information theoretic
  concepts. Neural Networks  \textbf{117},  104--123 (2019)

\end{thebibliography}

\fi

\end{document}